\documentclass[runningheads]{llncs}
\usepackage{url}
\usepackage{graphicx}
 \usepackage{nicematrix}

\usepackage{booktabs}
\usepackage{multirow}
\usepackage{color, colortbl}

\usepackage{pbox}
\usepackage{mathrsfs}
\usepackage[lined, ruled, linesnumbered, noend, scleft, nofillcomment]{algorithm2e}
\SetKwRepeat{Do}{do}{while}
\usepackage{amsmath,amsfonts}
\usepackage{algorithmic}
\usepackage{graphicx}
\usepackage{textcomp}

\usepackage{newtxmath}

\usepackage{caption}
\usepackage[utf8]{inputenc}
\usepackage{url}
\usepackage{amsmath}

\usepackage{siunitx}
\usepackage{svg}
\usepackage{parcolumns} 

\usepackage{scalerel,stackengine}
\stackMath
\newcommand\reallywidehat[1]{%
\savestack{\tmpbox}{\stretchto{%
  \scaleto{%
    \scalerel*[\widthof{\ensuremath{#1}}]{\kern-.6pt\bigwedge\kern-.6pt}%
    {\rule[-\textheight/2]{1ex}{\textheight}}
  }{\textheight}%
}{0.5ex}}%
\stackon[1pt]{#1}{\tmpbox}%
}
\begin{document}
\title{BOLIMES: Boruta–LIME optiMized fEature Selection for Gene Expression Classification}
\titlerunning{BOLIMES: Boruta–LIME optiMized fEature Selection for Gene Expression Classification}

%
\author{Bich-Chung Phan \inst{1}\and Thanh Ma\inst{1}\and Huu-Hoa Nguyen\inst{1} \and Thanh-Nghi Do\inst{1}}
\authorrunning{Bich-Chung Phan et al.}
%
\institute{Can Tho University, Can Tho, Vietnam\\
\email{\{pbchung,mtthanh,nhhoa,dtnghi\}@ctu.edu.vn}
}
\maketitle              
\begin{abstract}
Gene expression classification is a pivotal yet challenging task in bioinformatics, primarily due to the high dimensionality of genomic data and the risk of overfitting. To bridge this gap, we propose BOLIMES, a novel feature selection algorithm designed to enhance gene expression classification by systematically refining the feature subset. Unlike conventional methods that rely solely on statistical ranking or classifier-specific selection, we integrate the robustness of Boruta with the interpretability of LIME, ensuring that only the most relevant and influential genes are retained. BOLIMES first employs Boruta to filter out non-informative genes by comparing each feature against its randomized counterpart, thus preserving valuable information. It then uses LIME to rank the remaining genes based on their local importance to the classifier. Finally, an iterative classification evaluation determines the optimal feature subset by selecting the number of genes that maximizes predictive accuracy. By combining exhaustive feature selection with interpretability-driven refinement, our solution effectively balances dimensionality reduction with high classification performance, offering a powerful solution for high-dimensional gene expression analysis.

\end{abstract}
\section{Introduction}\label{sec1}

Gene expression classification \cite{ahmed2019gene,huynh2018random,do2024enhancing,do2023ensemble,huynh2019novel} has emerged as a fundamental tool in bioinformatics, enabling the identification of disease subtypes, prediction of patient outcomes, and discovery of potential therapeutic targets. With the advent of high-throughput sequencing technologies, researchers can now analyze vast gene expression profiles across thousands of genes simultaneously. However, this progress comes with a significant computational and analytical challenge: the \textit{curse of dimensionality} \cite{koppen2000curse,bach2017breaking}. In typical gene expression datasets, the number of genes (\( p \)) far exceeds the number of samples (\( n \)), often by several orders of magnitude. This imbalance leads to severe overfitting in machine learning models, where classifiers struggle to generalize due to the overwhelming presence of irrelevant or redundant features. Additionally, the high dimensionality increases computational complexity, making conventional classification models inefficient and impractical for real-world applications.

To address these challenges, \textit{feature selection} \cite{li2017feature,miao2016survey} plays a critical role in gene expression classification. By identifying and retaining only the most informative genes, feature selection not only improves model generalization but also enhances interpretability, allowing researchers to derive biologically meaningful insights from machine learning predictions \cite{do2024enhancing,karim2019onconetexplainer}. Furthermore, reducing the feature space significantly lowers computational costs, enabling faster model training and inference. More importantly, gene selection aligns with the biological reality that only a subset of genes actively contributes to disease mechanisms, making feature selection a crucial step in biomedical analysis. Without an effective selection strategy, classifiers are prone to noise, reduced accuracy, and difficulty in extracting meaningful biomarkers for clinical applications.

Given the importance of feature selection, extensive research has been conducted to develop robust selection techniques tailored for gene expression data. Traditional methods include filter-based approaches \cite{duch2006filter,cervante2012binary,cherrington2019feature,lee2011filter}, which rely on statistical measures such as mutual information, correlation, and entropy to rank features independently of the classifier. Although computationally efficient, filter methods often fail to capture complex gene interactions. Wrapper-based methods \cite{bajer2020wrapper,wang2014stability,kasongo2020deep,mufassirin2018novel}, such as recursive feature elimination (RFE) \cite{chen2007enhanced,yan2015feature} and genetic algorithms \cite{leardi1992genetic}, iteratively refine feature subsets based on classifier performance but tend to be computationally expensive for high-dimensional data. Meanwhile, embedded techniques \cite{imani2013novel,stanczyk2015feature,hamed2014accurate}, such as LASSO \cite{fonti2017feature,muthukrishnan2016lasso} and tree-based models \cite{freeman2013feature,liu2019embedded}, integrate feature selection within the classification process but may not always provide optimal feature subsets for different classifiers. Despite these advancements, no single approach consistently outperforms others across all gene expression datasets, necessitating hybrid strategies that leverage the strengths of multiple selection paradigms.

To bridge this gap, we propose BOLIMES (Boruta–LIME Enhanced eXtraction), a novel feature selection algorithm designed to enhance gene expression classification by systematically refining the feature subset. Unlike conventional methods that rely solely on statistical ranking or classifier-specific selection, BOLIMES integrates the robustness of Boruta \cite{kursa2010boruta,zhou2023diabetes} with the interpretability of LIME \cite{ribeiro2016should}, ensuring that only the most relevant and influential genes are retained. BOLIMES begins by applying Boruta, safeguarding against the premature elimination of any informative genes. Next, LIME (Local Interpretable Model-agnostic Explanations) \cite{ribeiro2016should,kumarakulasinghe2020evaluating} is employed to assess the local importance of each selected gene, offering an interpretable ranking that highlights the contributions of individual features to the model. Finally, an iterative classification evaluation identifies the optimal subset of features by selecting the number of genes that maximize predictive accuracy. We pay attention at Boruta and LIME for the following reasons: (1) Due to the large number of dimensions, LIME would face scalability issues when it comes to interpretation. (2) Boruta is viewed as a global feature selection method, whereas LIME focuses on local feature selection. Therefore, we will first perform global filtering using Boruta, followed by local refinement with LIME. (3) Our primary objective is to determine the optimal number of features for selection. As a result, we have proposed the Bolimex algorithm to select the most appropriate set of features, ensuring the best possible accuracy.

The remainder of this paper\footnote{This paper will be condensed to meet the page limit requirements of the conference upon acceptance. We aim to retain all the core ideas and details from this version to ensure the integrity of our research.} is structured as follows: we briefly presents a fundamental background in Section 2. Next, Section 3 describes our proposal algorithm. Then, Section 4 represents the experiment, the results of the models and an discussion. Finally, Section 5 shows the conclusions and future work.


\section{Background}
In this section, we present the foundation of our approach with integrating Boruta and Lime. We also provide the ML models that will be used for the gene expression classification task.
\subsection{Gene Expression Classification with ML models}
Gene expression classification \cite{do2024enhancing,do2023ensemble,huynh2019novel} lies at the forefront of biomedical research, offering profound insights into the molecular mechanisms underlying various diseases. ML models have become indispensable in this domain, as they can uncover complex patterns within vast and high-dimensional gene expression datasets. However, these datasets often contain a plethora of features, many of which are redundant or irrelevant, potentially obscuring the most critical biological signals and leading to overfitting. Consequently, feature selection becomes imperative—it refines the dataset by isolating the most informative genes, thereby enhancing model accuracy, interpretability, and computational efficiency. By focusing solely on the pivotal biomarkers, this research is able to achieve more reliable predictive outcomes. In this paper, we investigate and evaluate the classification with various ML techniques. Namely, we experiment our selected features with ML algorithms, i.e., SVM \cite{vapnik1995support}, Random Forest \cite{breiman2001random}, XGB \cite{chen2015xgboost}, Gradient Boosting \cite{friedman2002stochastic}.

\begin{definition}[Classification]
Let \( D = (X, y) \) be a dataset where \( X \subseteq \mathbb{R}^n \) is the feature space and \( y \in \mathcal{Y} = \{1,2,\dots,k\} \) represents the class labels. A classifier is a function
\[
f: X \to \mathcal{Y},
\]
that assigns a predicted label \( \hat{y} = f(x) \) to each input \( x \in X \). The function \( f \) is learned from the labeled examples
\[
D = \{(x_i, y_i) \mid x_i \in X,\; y_i \in \mathcal{Y},\; i = 1, \dots, N\},
\]
by minimizing a loss function \( \ell: \mathcal{Y} \times \mathcal{Y} \to \mathbb{R}_{\ge 0} \) that quantifies the error between the predicted and true labels. Once trained, \( f \) is used to classify new, unseen inputs.
\end{definition}


Feature selection is crucial before classification begins. Our study focuses on two techniques: Boruta and LIME. 
We now introduce Boruta and LIME in the following sections.

\subsection{Leveraging Boruta for Robust Feature Extraction}
Boruta \cite{kursa2010boruta,zhou2023diabetes} is a powerful wrapper-based feature selection algorithm designed to identify all truly relevant variables in a dataset. By comparing the importance of actual features with that of randomly generated ``shadow'' features, Boruta systematically filters out irrelevant variables while preserving essential predictors. This rigorous selection process is particularly valuable in high-dimensional applications, such as gene expression classification, where capturing meaningful signals is crucial. For clarity, we formally define Boruta as follows:
\begin{definition}[Boruta Feature Selection]
Let \( D = (X, y) \) be a dataset with features \( X = \{x_1, x_2, \dots, x_p\} \) and target \( y \). The Boruta algorithm identifies all relevant features in \( X \) as follows:
\begin{enumerate}
    \item \textbf{Shadow Feature Generation:} For each \( x_i \in X \), create a shadow feature \( x_i^{\text{shadow}} \) by randomly permuting its values, forming the set \( X^{\text{shadow}} \).
    \item \textbf{Importance Estimation:} Train a classifier (e.g., Random Forest) on the combined set \( X \cup X^{\text{shadow}} \) and compute the importance score \( I(z) \) for each \( z \).
    \item \textbf{Feature Comparison:} For each \( x_i \), define
    \[
    I^{\text{shadow}}_{\max} = \max_{z \in X^{\text{shadow}}} I(z).
    \]
    Then classify \( x_i \) as \emph{relevant} if \( I(x_i) \) is significantly greater than \( I^{\text{shadow}}_{\max} \), \emph{irrelevant} if significantly lower, or \emph{tentative} otherwise.
    \item \textbf{Iteration:} Remove irrelevant and tentative features and repeat until all features are decisively classified.
\end{enumerate}
The final selected subset \( X^* \subseteq X \) comprises all features deemed relevant.
\end{definition}

After applying the Boruta algorithm, we retain only the relevant features (confirmed) and excluded the tentative and irrelevant features (rejected). To further enhance the selection of features in \(X^*\), we employed the AI explanation technique outlined in the following section.


\subsection{XAI for Feature Selection}
Explainable AI (XAI) \cite{dwivedi2023explainable,zacharias2022designing} represents a forefront of AI research, aiming to elucidate the decision-making processes of complex models. In the context of gene expression classification, where feature selection is pivotal to model performance and interpretability, our study leverages LIME—Local Interpretable Model-Agnostic Explanations—to demystify and select critical features. LIME approximates the behavior of a sophisticated, black-box model with a simpler, locally interpretable surrogate, thereby pinpointing the most influential predictors in the vicinity of a given instance. This approach enhances the transparency of the model's predictions and facilitates a more informed and rigorous feature selection process, ultimately contributing to both improved accuracy and trustworthiness of the classification system.  Now, we provide a formal definition of LIME as follows:

\begin{definition}[LIME-based Feature Selection]
Let \( D^* = (X^*, y) \) be the dataset resulting from Boruta, where \( X^* \subseteq \mathbb{R}^{p^*} \) is the set of relevant features. Given a trained black-box classifier \( f: X^* \to \mathcal{Y} \) and an instance \( x \in X^* \), LIME constructs an interpretable surrogate model \( g \in G \) (typically linear), expressed as
\[
g(z) = w_0 + \sum_{j=1}^{p^*} w_j z_j,
\]
by solving the optimization problem
\[
\min_{g \in G} \sum_{z \in Z_x} \pi_x(z) \left( f(z) - g(z) \right)^2 + \Omega(g),
\]
where \( Z_x \) is a set of perturbed samples in the neighborhood of \( x \), \( \pi_x(z) \) is a proximity measure, and \( \Omega(g) \) enforces simplicity. The absolute coefficients \( |w_j| \) indicate the local importance of each feature, enabling a further refined selection \( X_{\text{selected}} \subseteq X^* \) for classification.
\end{definition}

To clarify, our choice of LIME for feature selection arises from the critical question of determining the optimal number of features for the model. In this context, assessing the local importance of each vector proves to be the most effective strategy, leading us to introduce the BOLIMES algorithm. The following section will provide a comprehensive explanation of the BOLIMES algorithm and its application.


\section{BOLIMES algorithm}\label{lb-ViFDFramwork}
\hspace{0.5cm}
Our methodology begins with applying the Boruta algorithm to sift through the high-dimensional gene expression dataset, effectively filtering out irrelevant features and isolating those that are truly significant. This initial reduction is critical because LIME, our subsequent interpretability tool, involves generating numerous perturbed samples and calculating distances—a computationally intensive process, especially in large feature spaces. By narrowing the feature set, we improve LIME's precision and enhance the model's interpretability, ultimately boosting classification performance.

However, the key question remains: how many features should be selected for optimal classification? In our proposed BOLIMES algorithm, which integrates Boruta and LIME, we first reduce the high-dimensional feature set by eliminating irrelevant variables with Boruta. Then, we further refine this subset using LIME to assess the local importance of each feature. Finally, we determine the optimal number of features by evaluating classification performance—selecting the subset that yields the highest accuracy for model training. In general, our model is both efficient and robust, relying only on the most informative features for gene expression classification.

\begin{algorithm}[h]
\scriptsize
    \DontPrintSemicolon
    \SetKwInOut{Input}{Input}
    \SetKwInOut{Output}{Output}
    
    \Input{Dataset \( D = (X, y) \) with \( X \subseteq \mathbb{R}^p \) and class labels \( y \).}
    \Output{Optimal feature subset \( X_{\text{opt}} \) and trained classifier \( f_{\text{opt}} \).}
    
    \Begin{
        \( X^* \leftarrow \text{Boruta}(D) \) \tcp*{\scriptsize Identify relevant features from \( X \)}
        
        \( \mathcal{I} \leftarrow \text{LIME}(f, X^*) \) \tcp*{\scriptsize Compute local importance scores on \( X^* \)}
        
        \( X^{*}_R \gets \{ x^*_{(i)} \}_{i=1}^{|X^*|} \) \quad  \( \mathcal{I}(x^*_{(1)}) \geq \dots \geq \mathcal{I}(x^*_{(|X^*|)}) \) \tcp*{\tiny \textbf{Rank} features in \( X^* \) in descending order of \( \mathcal{I} \)}
        
        \( \text{best\_acc} \leftarrow 0 \), \( k^* \leftarrow 0 \)\;
        
        \For{\( k = 10 \) \KwTo \( |X^*| \)}{
            \( S_k \gets \{ x^*_{(i)} \}_{i=1}^{k} \) \tcp*{\scriptsize Select top-\( k \) features from \( X^*_R \)}
            \( f_k \leftarrow \text{TrainClassifier}(D_{S_k}) \) \tcp*{\scriptsize Train classifier on \( D_{S_k} \)}
            \( \text{acc} \leftarrow \text{Evaluate}(f_k, D_{S_k}) \) \tcp*{\scriptsize Compute classification accuracy}
            
            \If{\( \text{acc} > \text{best\_acc} \)}{
                \( \text{best\_acc} \leftarrow \text{acc} \)\;
                \( k^* \leftarrow k \)\;
                \( f_{\text{opt}} \leftarrow f_k \)\;
            }
        }
        
        \( X_{\text{opt}} \gets \{ x^*_{(i)} \}_{i=1}^{k^*} \) \tcp*{\scriptsize Select top-\( k^* \) features from \( X^*_R \)}
        
        \Return{\( X_{\text{opt}}, f_{\text{opt}} \)}
    }
    
    \caption{Optimal Feature Selection for Classification using Boruta and LIME}
    \label{alg:BOLIMES}
\end{algorithm}


Algorithm \ref{alg:BOLIMES} presents an optimal feature selection framework by integrating Boruta and LIME to refine feature subsets for classification. Given a dataset \( D = (X, y) \), it first applies Boruta to extract a subset \( X^* \) of relevant features. LIME then computes local importance scores \( \mathcal{I} \) for \( X^* \), producing a ranked set \( X^*_R \) where
\[
X^*_R \gets \{ x^*_{(i)} \}_{i=1}^{|X^*|}, \quad \mathcal{I}(x^*_{(1)}) \geq \dots \geq \mathcal{I}(x^*_{(|X^*|)}).
\]
An iterative search determines the optimal number of features \( k^* \) by selecting the top-\( k \) ranked features, training a classifier \( f_k \), and evaluating its accuracy:
\[
S_k \gets \{ x^*_{(i)} \}_{i=1}^{k}, \quad f_k \leftarrow \text{TrainClassifier}(D_{S_k}).
\]
Note that we begin with 
$k=10$ and increase it until 
$|X^*|$, as using a smaller number of vectors would be inefficient. The best-performing classifier defines \( k^* \), yielding the final subset \( X_{\text{opt}} \) and trained model \( f_{\text{opt}} \).

The algorithm's complexity is driven by three key stages: Boruta for feature selection, LIME for ranking, and iterative classification. Boruta, relying on multiple iterations of Random Forest, has a worst-case complexity of \( O(T \cdot p^2 \log p) \). LIME, which perturbs \( m \) samples per feature, contributes \( O(m \cdot p) \). The final stage trains classifiers iteratively over \( p^* \) ranked features, leading to an overhead of \( O(n p^*{}^2) \) assuming a model with \( O(n k) \) complexity. Thus, the total complexity is:
\(
O(T \cdot p^2 \log p) + O(m \cdot p) + O(n p^*{}^2),
\)
where \( p^* \ll p \) in practice, making the approach feasible for moderate-dimensional data but computationally intensive for extremely large \( p \).



\section{Experiment and results}
\label{lb-experimentalresult}
\hspace{0.5cm}
This section offers a brief description of the gene expression datasets while delivering a detailed comparative analysis of the classification models. Furthermore, we also provide the results of Boruta. Our source code has been made publicly accessible on GitHub\footnote{\url{https://github.com/Pbchung75/BOLIMES/}}.

 \subsection{Dataset and Configurations}
We carry out a series of experiments using 14 gene expression datasets, all meticulously sourced from the reputable ArrayExpress repository \cite{brazma2003arrayexpress}. The gene expression datasets \cite{do2024enhancing} summarized in Table~\ref{tab:datasets} exemplify the inherent challenges of high-dimensional biomedical data. With sample sizes ranging from $53$ to $575$ and feature counts spanning from approximately $11,950$ to over $54,600$, these datasets present a significant imbalance between the number of available samples and the vast dimensionality of gene expression profiles. Additionally, the variability in the number of classes—from as few as $3$ to as many as $10$—further complicates the classification task by introducing diverse and complex biological signals. 
This high dimensionality coupled with limited sample sizes accentuates the risk of overfitting and underscores the critical need for effective feature selection. 
In this study, our feature selection strategy is specifically designed to address these challenges, ensuring that only the most relevant features are retained for subsequent classification tasks.

\begin{table}[ht]
    \centering
    \caption{Dataset Characteristics}
    \label{tab:datasets}
    \begin{tabular}{cccccc}
        \toprule
        ID & Dataset        & \#Datapoints & \#Dimensions & \#Classes \\
        \midrule
        1  & E-GEOD-20685   & 327          & 54627        & 6  \\
        2  & E-GEOD-20711   & 90           & 54675        & 5  \\
        3  & E-GEOD-21050   & 310          & 54613        & 4  \\
        4  & E-GEOD-21122   & 158          & 22283        & 7  \\
        5  & E-GEOD-29354   & 53           & 22215        & 3  \\
        6  & E-GEOD-30784   & 229          & 54675        & 3  \\
        7  & E-GEOD-31312   & 498          & 54630        & 3  \\
        8  & E-GEOD-31552   & 111          & 33297        & 3  \\
        9  & E-GEOD-32537   & 217          & 11950        & 7  \\
        10 & E-GEOD-33315   & 575          & 22283        & 10 \\
        11 & E-GEOD-37364   & 94           & 54675        & 4  \\
        12 & E-GEOD-39582   & 566          & 54675        & 6  \\
        13 & E-GEOD-39716   & 53           & 33297        & 3  \\
        14 & E-GEOD-44077   & 226          & 33252        & 4  \\
        \bottomrule
    \end{tabular}
\end{table}

To evaluate our approach and the training model, we implement the AI libraries \textit{(i.e., Pandas (version 2.2.2),
Scikit-learn (version 1.6.1))} and run the experiments on the computer with the following configuration: \textit{Intel Core i5-12400, 2.50 GHz, 24 GB RAM, Windows 11 Pro OS}.

 \subsection{Boruta's Feature Selection}
In this study, we adopt Boruta as our primary feature selection method to substantially reduce the dimensionality of our gene expression datasets for the \textit{global} problem. The rationale behind this choice is to prevent an explosion in computational complexity and potential loss of interpretability when LIME is applied to an excessively high-dimensional feature space. By using Boruta, we are able to effectively eliminate irrelevant features (tentative and rejected in Table~\ref{tab:feature_selection_results}) while retaining those that are truly informative for classification. As evidenced in Table~\ref{tab:feature_selection_results}, datasets such as \textit{E-GEOD-20685} are reduced from 54,627 dimensions to only 545 confirmed features, thereby rendering the subsequent LIME analysis both feasible and efficient. To achieve this, Boruta is executed with the following parameters via the BorutaPy library: a random forest classifier (`rf`) with 300 estimators, a maximum of 200 iterations, an alpha value of $0.01$, a strict percentile threshold of 100, two-step feature selection enabled, and a fixed random state of 42, with verbose output enabled. These parameter settings are meticulously selected to ensure a rigorous and robust selection process, ultimately facilitating a more interpretable and high-performing classification model.

\begin{table}[ht]
    \centering
    \caption{Feature Selection Results}
    \label{tab:feature_selection_results}
    \begin{tabular}{cccccc}
        \toprule
        ID & Dataset & & Features & & Feature Selection \\
        & E-GEOD-* & Confirmed & Tentative  & Rejected & Time (s) \\
        \midrule
        1  & 20685 & 545  & 262 & 53820 & 633,0205021 \\
        2  & 20711 & 111  & 33  & 54531 & 140,0120337 \\
        3  & 21050 & 72   & 17  & 54524 & 350,3323104 \\
        4  & 21122 & 271  & 124 & 21888 & 219,2371044 \\
        5  & 29354 & 28   & 3   & 22184 & 91,59759808 \\
        6  & 30784 & 171  & 64  & 54440 & 279,9282582 \\
        7  & 31312 & 213  & 42  & 54375 & 777,7057292 \\
        8  & 31552 & 79   & 48  & 33170 & 136,9728532 \\
        9  & 32537 & 96   & 17  & 11837 & 232,6542134 \\
        10 & 33315 & 483  & 108 & 21692 & 1176,811348  \\
        11 & 37364 & 59   & 43  & 54573 & 129,0980539  \\
        12 & 39582 & 640  & 253 & 53782 & 1306,720685  \\
        13 & 39716 & 124  & 29  & 33144 & 99,24336505  \\
        14 & 44077 & 227  & 231 & 32794 & 221,3678348  \\
        \bottomrule
    \end{tabular}
\end{table}

 \subsection{Classification results}
In this study, we conduct experiments with four ML algorithms (using SVM, Random Forest (RF), XGBoost (XGB), and Gradient Boosting (GB)) to determine the method yielding the highest accuracy across various gene expression datasets (see Table \ref{tab:svm_results}, \ref{tab:RF_results}, \ref{tab:XGB_results}, and \ref{tab:gb_results}). We configure an \textit{SVM} using SVC with a radial basis function kernel, setting C to $100,000$, gamma to $0.001$, ensuring reproducibility with a random state of $42$. In parallel, we employe a Random Forest classifier with $200$ estimators, a maximum depth of $10$. Additionally, we implemented both an \textit{XGBClassifier} and a \textit{GradientBoostingClassifier}, each configured with $50$ estimators, a maximum depth of $10$, and a learning rate of $0.01$. Our comparative evaluation reveales that while the SVM excelled on datasets with fewer samples, the ensemble methods—particularly Random Forest and XGBoost—demonstrated more robust performance on high-dimensional data. 

In Table \ref{tab:svm_results}, \ref{tab:RF_results}, \ref{tab:XGB_results}, and \ref{tab:gb_results}, we conduct SVM, RF, and Decision Tree (DT) methodology to identify the importance of features and the number of selected features ($k$ selected dimensions) following the methodology outlined in \cite{do2024enhancing} i.e., Top $k$ selected feature of E-GEOD-20685 (ID 1) is $85$, E-GEOD-20711 (ID 2) is $43$, E-GEOD-21050 (ID 3) is $275$, and others. Overall, our proposed algorithm demonstrated superior performance, achieving higher accuracy with respect to the F1-score. In certain instances, it reached $100\%$, while other solutions (SVM, RF, DT) only achieved $80-90\%$. The feature selection algorithms, such as SVM, RF, and DT, with a smaller number of features, appeared to underperform in covering the data (e.g., IDs 1, 2, 4). Our approach, however, seemed to be more effective in determining the optimal number of features needed.

\begin{table}[htbp]
  \centering
  \caption{SVM Classification Results with Top $k$ Features}
  \label{tab:svm_results}
  \scriptsize
  \begin{tabular}{cccccccccccc}
    \toprule
    ID & Dataset & Class & Method & Samples & \multicolumn{1}{c}{Top k} & \multicolumn{4}{c}{Classification Results} & {Training} \\
    \cmidrule(lr){7-10} 
        &         &       &         &   & Features & Acc & Prec & Rec & F1 Score & Time (s)  \\
  \midrule
        1  & 20685 & 6 & ours & 327 & 107 & 0.955 & 0.956 & 0.955 & 0.955 & 112.137 \\
           &       &   & SVM  &     & 85  & 0.859 & 0.872 & 0.859 & 0.859 & 0.135 \\
           &       &   & RF   &     & 85  & 0.890 & 0.902 & 0.890 & 0.889 & 0.147 \\
           &       &   & DT   &     & 85  & 0.872 & 0.884 & 0.872 & 0.868 & 0.192 \\
        \midrule
        2  & 20711 & 5 & ours & 90  & 67  & 0.944 & 0.952 & 0.944 & 0.943 & 17.754 \\
           &       &   & SVM  &     & 43  & 0.811 & 1.000 & 0.811 & 0.811 & 0.818 \\
           &       &   & RF   &     & 43  & 0.778 & 1.000 & 0.778 & 0.778 & 0.806 \\
           &       &   & DT   &     & 43  & 0.778 & 1.000 & 0.778 & 0.778 & 0.788 \\
        \midrule
        3  & 21050 & 4 & ours & 310 & 70  & 0.677 & 0.709 & 0.677 & 0.678 & 295.982\\
           &       &   & SVM  &     & 275 & 0.684 & 0.690 & 0.684 & 0.676 & 0.218 \\
           &       &   & RF   &     & 275 & 0.681 & 0.681 & 0.681 & 0.666 & 0.227 \\
           &       &   & DT   &     & 275 & 0.652 & 0.674 & 0.652 & 0.622 & 0.250 \\
        \midrule
        4  & 21122 & 7 & ours & 158 & 170 & 0.938 & 0.915 & 0.938 & 0.922 & 82.092 \\
           &       &   & SVM  &     & 78  & 0.886 & 1.000 & 0.886 & 0.886 & 1.489 \\
           &       &   & RF   &     & 78  & 0.880 & 1.000 & 0.880 & 0.880 & 1.483 \\
           &       &   & DT   &     & 78  & 0.842 & 1.000 & 0.842 & 0.842 & 1.744 \\
        \midrule
        5  & 29354 & 3 & ours & 53  & 10  & 1.000 & 1.000 & 1.000 & 1.000 & 0.165 \\
           &       &   & SVM  &     & 35  & 0.868 & 1.000 & 0.868 & 0.868 & 0.415 \\
           &       &   & RF   &     & 35  & 0.830 & 1.000 & 0.830 & 0.830 & 0.422 \\
           &       &   & DT   &     & 35  & 0.660 & 1.000 & 0.660 & 0.660 & 0.397 \\
        \midrule
        6  & 30784 & 3 & ours & 229 & 14  & 1.000 & 1.000 & 1.000 & 1.000 & 3.476 \\
           &       &   & SVM  &     & 42  & 0.904 & 1.000 & 0.904 & 0.904 & 2.277 \\
           &       &   & RF   &     & 42  & 0.908 & 1.000 & 0.908 & 0.908 & 2.075 \\
           &       &   & DT   &     & 42  & 0.908 & 1.000 & 0.908 & 0.908 & 2.175 \\
        \midrule
        7  & 31312 & 3 & ours & 498 & 43  & 0.900 & 0.900 & 0.900 & 0.900 & 2.144 \\
           &       &   & SVM  &     & 195 & 0.878 & 0.872 & 0.878 & 0.859 & 0.236 \\
           &       &   & RF   &     & 195 & 0.878 & 0.864 & 0.878 & 0.857 & 0.225 \\
           &       &   & DT   &     & 195 & 0.809 & 0.796 & 0.809 & 0.775 & 0.345 \\
\midrule
        8  & 31552 & 3 & ours & 111 & 15  & 0.957 & 0.917 & 0.957 & 0.936 & 1.816 \\
           &       &   & SVM  &     & 44  & 0.820 & 1.000 & 0.820 & 0.820 & 0.931 \\
           &       &   & RF   &     & 44  & 0.883 & 1.000 & 0.883 & 0.883 & 0.870 \\
           &       &   & DT   &     & 44  & 0.847 & 1.000 & 0.847 & 0.847 & 0.890 \\
        \midrule
        9  & 32537 & 7 & ours & 217 & 40  & 0.795 & 0.730 & 0.795 & 0.747 & 21.201 \\
           &       &   & SVM  &     & 171 & 0.760 & 1.000 & 0.760 & 0.760 & 3.054 \\
           &       &   & RF   &     & 171 & 0.779 & 1.000 & 0.779 & 0.779 & 2.910 \\
           &       &   & DT   &     & 171 & 0.756 & 1.000 & 0.756 & 0.756 & 3.450 \\
        \midrule
        10 & 33315 & 10 & ours & 575 & 273 & 0.878 & 0.860 & 0.878 & 0.867 & 33.250 \\
           &       &    & SVM  &     & 2321 & 0.873 & 0.893 & 0.873 & 0.855 & 4.174 \\
           &       &    & RF   &     & 2321 & 0.870 & 0.893 & 0.870 & 0.851 & 4.667 \\
           &       &    & DT   &     & 2321 & 0.781 & 0.845 & 0.781 & 0.752 & 5.534 \\
        \midrule
        11 & 37364 & 4 & ours & 94  & 17  & 0.947 & 0.965 & 0.947 & 0.947 & 2.083 \\
           &       &   & SVM  &     & 140 & 0.809 & 1.000 & 0.809 & 0.809 & 0.793 \\
           &       &   & RF   &     & 140 & 0.755 & 1.000 & 0.755 & 0.755 & 0.750 \\
           &       &   & DT   &     & 140 & 0.713 & 1.000 & 0.713 & 0.713 & 0.798 \\
        \midrule
        12 & 39582 & 6 & ours & 566 & 251 & 0.877 & 0.889 & 0.877 & 0.880 & 25.058\\
           &       &   & SVM  &     & 441 & 0.859 & 0.873 & 0.859 & 0.859 & 0.466 \\
           &       &   & RF   &     & 441 & 0.862 & 0.874 & 0.862 & 0.861 & 0.480 \\
           &       &   & DT   &     & 441 & 0.790 & 0.812 & 0.790 & 0.786 & 0.764 \\
        \midrule
        13 & 39716 & 3 & ours & 53  & 12  & 1.000 & 1.000 & 1.000 & 1.000 & 0.454 \\
           &       &   & SVM  &     & 118 & 0.887 & 1.000 & 0.887 & 0.887 & 0.408 \\
           &       &   & RF   &     & 118 & 0.925 & 1.000 & 0.925 & 0.925 & 0.408 \\
           &       &   & DT   &     & 118 & 0.774 & 1.000 & 0.774 & 0.774 & 0.401 \\
        \midrule
        14 & 44077 & 4 & ours & 226 & 18  & 1.000 & 1.000 & 1.000 & 1.000 & 5.220 \\
           &       &   & SVM  &     & 23  & 0.991 & 1.000 & 0.991 & 0.991 & 1.803 \\
           &       &   & RF   &     & 23  & 0.996 & 1.000 & 0.996 & 0.996 & 1.790 \\
           &       &   & DT   &     & 23  & 0.987 & 1.000 & 0.987 & 0.987 & 2.005 \\
    
    \midrule
  \end{tabular}
\end{table}



\begin{table}[htbp]
  \centering
  \caption{Random Forest Classification Results with Top $k$ Features}
  \label{tab:RF_results}
  \scriptsize
  \begin{tabular}{cccccccccccc}
    \toprule
    ID & Dataset & Class & Method & Samples & \multicolumn{1}{c}{Top k} & \multicolumn{4}{c}{Classification Results} & {Training} \\
    \cmidrule(lr){7-10} 
        &         &       &         &   & Features & Acc & Prec & Rec & F1 Score & Time (s)  \\
  \midrule
        1  & 20685 & 6 & Ours & 327 & 48 & 0.970 & 0.971 & 0.970 & 0.970 & 16306.229 \\
           &       &   & SVM  &     & 85  & 0.838 & 0.852 & 0.838 & 0.831 & 500.484 \\
           &       &   & RF   &     & 85  & 0.874 & 0.890 & 0.874 & 0.872 & 4.706 \\
           &       &   & DT   &     & 85  & 0.887 & 0.904 & 0.887 & 0.885 & 5.175 \\
        \midrule
        2  & 20711 & 5 & Ours & 90  & 10  & 0.833 & 0.841 & 0.833 & 0.830 & 5.396 \\
           &       &   & SVM  &     & 43  & 0.767 & 1.000 & 0.767 & 0.767 & 25.742 \\
           &       &   & RF   &     & 43  & 0.833 & 1.000 & 0.833 & 0.833 & 2593.662 \\
           &       &   & DT   &     & 43  & 0.833 & 1.000 & 0.833 & 0.833 & 2588.751 \\
        \midrule
        3  & 21050 & 4 & Ours & 310 & 72  & 0.726 & 0.727 & 0.726 & 0.711 & 1507.624 \\
           &       &   & SVM  &     & 275 & 0.732 & 0.776 & 0.732 & 0.696 & 6.371 \\
           &       &   & RF   &     & 275 & 0.735 & 0.773 & 0.735 & 0.700 & 6.617 \\
           &       &   & DT   &     & 275 & 0.739 & 0.793 & 0.739 & 0.701 & 6.441 \\
        \midrule
        4  & 21122 & 7 & Ours & 158 & 104 & 0.938 & 0.922 & 0.938 & 0.925 & 1039.209 \\
           &       &   & SVM  &     & 78  & 0.880 & 1.000 & 0.880 & 0.880 & 50.215 \\
           &       &   & RF   &     & 78  & 0.886 & 1.000 & 0.886 & 0.886 & 50.446 \\
           &       &   & DT   &     & 78  & 0.873 & 1.000 & 0.873 & 0.873 & 53.083 \\
        \midrule
        5  & 29354 & 3 & Ours & 53  & 10  & 1.000 & 1.000 & 1.000 & 1.000 & 3.325 \\
           &       &   & SVM  &     & 35  & 0.887 & 1.000 & 0.887 & 0.887 & 13.873 \\
           &       &   & RF   &     & 35  & 0.811 & 1.000 & 0.811 & 0.811 & 12.891 \\
           &       &   & DT   &     & 35  & 0.774 & 1.000 & 0.774 & 0.774 & 12.654 \\
        \midrule
        6  & 30784 & 3 & Ours & 229 & 12  & 1.000 & 1.000 & 1.000 & 1.000 & 43.917 \\
           &       &   & SVM  &     & 42  & 0.913 & 1.000 & 0.913 & 0.913 & 80.499 \\
           &       &   & RF   &     & 42  & 0.895 & 1.000 & 0.895 & 0.895 & 75.487 \\
           &       &   & DT   &     & 42  & 0.908 & 1.000 & 0.908 & 0.908 & 76.522 \\
        \midrule
        7  & 31312 & 3 & Ours & 498 & 161 & 0.900 & 0.821 & 0.900 & 0.858 & 221.322 \\
           &       &   & SVM  &     & 195 & 0.871 & 0.885 & 0.871 & 0.828 & 8.221 \\
           &       &   & RF   &     & 195 & 0.872 & 0.885 & 0.872 & 0.828 & 7.968 \\
           &       &   & DT   &     & 195 & 0.841 & 0.859 & 0.841 & 0.800 & 8.715 \\
     \midrule
        8  & 31552 & 3 & Ours & 111 & 12 & 1.000 & 1.000 & 1.000 & 1.000 & 20.318 \\
           &       &   & SVM  &     & 44  & 0.865 & 1.000 & 0.865 & 0.865 & 31.843 \\
           &       &   & RF   &     & 44  & 0.901 & 1.000 & 0.901 & 0.901 & 29.352 \\
           &       &   & DT   &     & 44  & 0.865 & 1.000 & 0.865 & 0.865 & 30.185 \\
        \midrule
        9  & 32537 & 7 & Ours & 217 & 87  & 0.773 & 0.679 & 0.773 & 0.707 & 1238.228 \\
           &       &   & SVM  &     & 171 & 0.779 & 1.000 & 0.779 & 0.779 & 113.423 \\
           &       &   & RF   &     & 171 & 0.793 & 1.000 & 0.793 & 0.793 & 104.033 \\
           &       &   & DT   &     & 171 & 0.765 & 1.000 & 0.765 & 0.765 & 110.804 \\
        \midrule
        10 & 33315 & 10 & Ours & 575 & 444 & 0.887 & 0.854 & 0.887 & 0.868 & 985.966 \\
           &       &    & SVM  &     & 2321 & 0.835 & 0.866 & 0.835 & 0.808 & 31.634 \\
           &       &    & RF   &     & 2321 & 0.833 & 0.868 & 0.833 & 0.800 & 31.258 \\
           &       &    & DT   &     & 2321 & 0.797 & 0.840 & 0.797 & 0.756 & 31.688 \\
        \midrule
        11 & 37364 & 4 & Ours & 94  & 10  & 0.947 & 0.953 & 0.947 & 0.947 & 5.579 \\
           &       &   & SVM  &     & 140 & 0.798 & 1.000 & 0.798 & 0.798 & 26.410 \\
           &       &   & RF   &     & 140 & 0.809 & 1.000 & 0.809 & 0.809 & 25.745 \\
           &       &   & DT   &     & 140 & 0.734 & 1.000 & 0.734 & 0.734 & 27.083 \\
        \midrule
        12 & 39582 & 6 & Ours & 566 & 189 & 0.877 & 0.890 & 0.877 & 0.876 & 304.137 \\
           &       &   & SVM  &     & 441 & 0.820 & 0.836 & 0.820 & 0.819 & 13.926 \\
           &       &   & RF   &     & 441 & 0.832 & 0.847 & 0.832 & 0.831 & 13.629 \\
           &       &   & DT   &     & 441 & 0.807 & 0.831 & 0.807 & 0.805 & 14.261 \\
        \midrule
        13 & 39716 & 3 & Ours & 53  & 15  & 1.000 & 1.000 & 1.000 & 1.000 & 19.338 \\
           &       &   & SVM  &     & 118 & 0.925 & 1.000 & 0.925 & 0.925 & 13.133 \\
           &       &   & RF   &     & 118 & 0.925 & 1.000 & 0.925 & 0.925 & 13.049 \\
           &       &   & DT   &     & 118 & 0.830 & 1.000 & 0.830 & 0.830 & 13.391 \\
        \midrule
        14 & 44077 & 4 & Ours & 226 & 52  & 1.000 & 1.000 & 1.000 & 1.000 & 642.643 \\
           &       &   & SVM  &     & 23  & 0.991 & 1.000 & 0.991 & 0.991 & 61.789 \\
           &       &   & RF   &     & 23  & 0.996 & 1.000 & 0.996 & 0.996 & 60.206 \\
           &       &   & DT   &     & 23  & 0.996 & 1.000 & 0.996 & 0.996 & 63.950 \\
    
    \bottomrule
  \end{tabular}
\end{table}

\begin{table}[htbp]
  \centering
  \caption{XGB Classification Results with Top $k$ Features}
  \label{tab:XGB_results}
  \scriptsize
  \begin{tabular}{cccccccccccc}
    \toprule
    ID & Dataset & Class & Method & Samples & \multicolumn{1}{c}{Top k} & \multicolumn{4}{c}{Classification Results} & {Training} \\
    \cmidrule(lr){7-10} 
        &         &       &         &   & Features & Acc & Prec & Rec & F1 Score & Time (s)  \\
  \midrule
        1  & 20685 & 6 & Ours & 327 & 114 & 0.955 & 0.961 & 0.955 & 0.955 & 40.015 \\
           &       &   & SVM  &     & 85  & 0.804 & 0.839 & 0.804 & 0.793 & 10.932 \\
           &       &   & RF   &     & 85  & 0.829 & 0.854 & 0.829 & 0.822 & 11.592 \\
           &       &   & DT   &     & 85  & 0.851 & 0.874 & 0.851 & 0.843 & 11.531 \\
        \midrule
        2  & 20711 & 5 & Ours & 90  & 10  & 0.833 & 0.841 & 0.833 & 0.830 & 0.328 \\
           &       &   & SVM  &     & 43  & 0.744 & 1.000 & 0.744 & 0.744 & 62.459 \\
           &       &   & RF   &     & 43  & 0.811 & 1.000 & 0.811 & 0.811 & 63.819 \\
           &       &   & DT   &     & 43  & 0.811 & 1.000 & 0.811 & 0.811 & 63.947 \\
        \midrule
        3  & 21050 & 4 & Ours & 310 & 60  & 0.790 & 0.791 & 0.790 & 0.784 & 16.183 \\
           &       &   & SVM  &     & 275 & 0.732 & 0.769 & 0.732 & 0.701 & 9.478 \\
           &       &   & RF   &     & 275 & 0.748 & 0.782 & 0.748 & 0.724 & 9.977 \\
           &       &   & DT   &     & 275 & 0.752 & 0.796 & 0.752 & 0.720 & 9.622 \\
        \midrule
        4  & 21122 & 7 & Ours & 158 & 159 & 0.938 & 0.921 & 0.938 & 0.924 & 44.087 \\
           &       &   & SVM  &     & 78  & 0.848 & 1.000 & 0.848 & 0.848 & 162.044 \\
           &       &   & RF   &     & 78  & 0.854 & 1.000 & 0.854 & 0.854 & 164.271 \\
           &       &   & DT   &     & 78  & 0.867 & 1.000 & 0.867 & 0.867 & 166.522 \\
        \midrule
        5  & 29354 & 3 & Ours & 53  & 10  & 1.000 & 1.000 & 1.000 & 1.000 & 0.271 \\
           &       &   & SVM  &     & 35  & 0.906 & 1.000 & 0.906 & 0.906 & 23.871 \\
           &       &   & RF   &     & 35  & 0.774 & 1.000 & 0.774 & 0.774 & 23.896 \\
           &       &   & DT   &     & 35  & 0.830 & 1.000 & 0.830 & 0.830 & 23.359 \\
        \midrule
        6  & 30784 & 3 & Ours & 229 & 50  & 0.978 & 0.979 & 0.978 & 0.976 & 6.344 \\
           &       &   & SVM  &     & 42  & 0.921 & 1.000 & 0.921 & 0.921 & 121.816 \\
           &       &   & RF   &     & 42  & 0.913 & 1.000 & 0.913 & 0.913 & 119.124 \\
           &       &   & DT   &     & 42  & 0.921 & 1.000 & 0.921 & 0.921 & 123.405 \\
        \midrule
        7  & 31312 & 3 & Ours & 498 & 10  & 0.860 & 0.848 & 0.860 & 0.843 & 0.388 \\
           &       &   & SVM  &     & 195 & 0.851 & 0.863 & 0.851 & 0.810 & 7.677 \\
           &       &   & RF   &     & 195 & 0.833 & 0.828 & 0.833 & 0.797 & 7.432 \\
           &       &   & DT   &     & 195 & 0.831 & 0.850 & 0.831 & 0.790 & 7.762 \\
        \midrule
        8  & 31552 & 3 & ours & 111 & 10  & 1.000 & 1.000 & 1.000 & 1.000 & 0.557 \\
           &       &   & SVM  &     & 44  & 0.856 & 1.000 & 0.856 & 0.856 & 53.782 \\
           &       &   & RF   &     & 44  & 0.892 & 1.000 & 0.892 & 0.892 & 52.092 \\
           &       &   & DT   &     & 44  & 0.892 & 1.000 & 0.892 & 0.892 & 53.220 \\
        \midrule
        9  & 32537 & 7 & ours & 217 & 95  & 0.795 & 0.693 & 0.795 & 0.736 & 26.388 \\
           &       &   & SVM  &     & 171 & 0.783 & 1.000 & 0.783 & 0.783 & 282.386 \\
           &       &   & RF   &     & 171 & 0.802 & 1.000 & 0.802 & 0.802 & 283.478 \\
           &       &   & DT   &     & 171 & 0.770 & 1.000 & 0.770 & 0.770 & 283.282 \\
        \midrule
        10 & 33315 & 10 & ours & 575 & 425 & 0.870 & 0.842 & 0.870 & 0.854 & 389.225 \\
           &       &    & SVM  &     & 2321 & 0.861 & 0.883 & 0.861 & 0.838 & 94.119 \\
           &       &    & RF   &     & 2321 & 0.859 & 0.879 & 0.859 & 0.839 & 94.236 \\
           &       &    & DT   &     & 2321 & 0.852 & 0.874 & 0.852 & 0.827 & 96.417 \\
        \midrule
        11 & 37364 & 4 & ours & 94  & 21  & 1.000 & 1.000 & 1.000 & 1.000 & 1.862 \\
           &       &   & SVM  &     & 140 & 0.777 & 1.000 & 0.777 & 0.777 & 59.093 \\
           &       &   & RF   &     & 140 & 0.798 & 1.000 & 0.798 & 0.798 & 59.041 \\
           &       &   & DT   &     & 140 & 0.745 & 1.000 & 0.745 & 0.745 & 60.416 \\
        \midrule
        12 & 39582 & 6 & ours & 566 & 556 & 0.825 & 0.833 & 0.825 & 0.824 & 554.762 \\
           &       &   & SVM  &     & 441 & 0.797 & 0.810 & 0.797 & 0.793 & 21.655 \\
           &       &   & RF   &     & 441 & 0.804 & 0.816 & 0.804 & 0.799 & 22.184 \\
           &       &   & DT   &     & 441 & 0.788 & 0.791 & 0.788 & 0.784 & 22.569 \\
        \midrule
        13 & 39716 & 3 & ours & 53  & 44  & 1.000 & 1.000 & 1.000 & 1.000 & 3.746 \\
           &       &   & SVM  &     & 118 & 0.897 & 1.000 & 0.887 & 0.887 & 23.631 \\
           &       &   & RF   &     & 118 & 0.925 & 1.000 & 0.925 & 0.925 & 22.742 \\
           &       &   & DT   &     & 118 & 0.943 & 1.000 & 0.943 & 0.943 & 24.227 \\
        \midrule
        14 & 44077 & 4 & ours & 226 & 57  & 0.978 & 0.980 & 0.978 & 0.978 & 8.332 \\
           &       &   & SVM  &     & 23  & 0.978 & 1.000 & 0.978 & 0.978 & 132.205 \\
           &       &   & RF   &     & 23  & 0.987 & 1.000 & 0.987 & 0.987 & 133.960 \\
           &       &   & DT   &     & 23  & 0.991 & 1.000 & 0.991 & 0.991 & 133.546 \\

    \bottomrule
  \end{tabular}
\end{table}

\begin{table}[htbp]
  \centering
  \caption{Gradient Boosting Classification Results with Top $k$ Features}
  \label{tab:gb_results}
  \scriptsize
  \begin{tabular}{cccccccccccc}
    \toprule
    ID & Dataset & Class & Method & Samples & \multicolumn{1}{c}{Top k} & \multicolumn{4}{c}{Classification Results} & {Training} \\
    \cmidrule(lr){7-10} 
        &         &       &         &   & Features & Acc & Prec & Rec & F1 Score & Time (s)  \\
  \midrule
        1  & 20685 & 6 & ours & 327 & 468 & 0.939 & 0.947 & 0.939 & 0.936 & 17.957 \\
           &       &   & SVM  &     & 85  & 0.795 & 0.820 & 0.795 & 0.788 & 53.840 \\
           &       &   & RF   &     & 85  & 0.856 & 0.876 & 0.856 & 0.855 & 53.966 \\
           &       &   & DT   &     & 85  & 0.844 & 0.865 & 0.844 & 0.834 & 53.381 \\
        \midrule
        2  & 20711 & 5 & ours & 90  & 38  & 0.889 & 0.921 & 0.889 & 0.892 & 0.664 \\
           &       &   & SVM  &     & 43  & 0.733 & 1.000 & 0.733 & 0.733 & 91.394 \\
           &       &   & RF   &     & 43  & 0.711 & 1.000 & 0.711 & 0.711 & 90.162 \\
           &       &   & DT   &     & 43  & 0.711 & 1.000 & 0.711 & 0.711 & 89.505 \\
        \midrule
        3  & 21050 & 4 & ours & 310 & 65  & 0.806 & 0.802 & 0.806 & 0.804 & 1.844 \\
           &       &   & SVM  &     & 275 & 0.703 & 0.733 & 0.703 & 0.685 & 72.745 \\
           &       &   & RF   &     & 275 & 0.735 & 0.745 & 0.735 & 0.720 & 78.348 \\
           &       &   & DT   &     & 275 & 0.758 & 0.792 & 0.758 & 0.742 & 76.235 \\
        \midrule
        4  & 21122 & 7 & ours & 158 & 257 & 0.875 & 0.872 & 0.875 & 0.871 & 5.544 \\
           &       &   & SVM  &     & 78  & 0.804 & 1.000 & 0.804 & 0.804 & 511.805 \\
           &       &   & RF   &     & 78  & 0.835 & 1.000 & 0.835 & 0.835 & 507.856 \\
           &       &   & DT   &     & 78  & 0.861 & 1.000 & 0.861 & 0.861 & 513.399 \\
        \midrule
        5  & 29354 & 3 & ours & 53  & 12  & 1.000 & 1.000 & 1.000 & 1.000 & 0.387 \\
           &       &   & SVM  &     & 35  & 0.849 & 1.000 & 0.849 & 0.849 & 22.807 \\
           &       &   & RF   &     & 35  & 0.717 & 1.000 & 0.717 & 0.717 & 20.737 \\
           &       &   & DT   &     & 35  & 0.811 & 1.000 & 0.811 & 0.811 & 20.797 \\
        \midrule
        6  & 30784 & 3 & ours & 229 & 39  & 0.957 & 0.957 & 0.957 & 0.957 & 0.747 \\
           &       &   & SVM  &     & 42  & 0.913 & 1.000 & 0.913 & 0.913 & 272.929 \\
           &       &   & RF   &     & 42  & 0.900 & 1.000 & 0.900 & 0.900 & 269.238 \\
           &       &   & DT   &     & 42  & 0.943 & 1.000 & 0.943 & 0.943 & 273.324 \\
        \midrule
        7  & 31312 & 3 & ours & 498 & 169 & 0.880 & 0.853 & 0.880 & 0.854 & 4.961 \\
           &       &   & SVM  &     & 195 & 0.853 & 0.840 & 0.853 & 0.830 & 89.338 \\
           &       &   & RF   &     & 195 & 0.853 & 0.828 & 0.853 & 0.824 & 90.304 \\
           &       &   & DT   &     & 195 & 0.825 & 0.805 & 0.825 & 0.795 & 90.659 \\
       \midrule
        8  & 31552 & 3 & ours & 111 & 13  & 0.957 & 0.917 & 0.957 & 0.936 & 0.310 \\
           &       &   & SVM  &     & 44  & 0.802 & 1.000 & 0.802 & 0.802 & 77.917 \\
           &       &   & RF   &     & 44  & 0.883 & 1.000 & 0.883 & 0.883 & 74.657 \\
           &       &   & DT   &     & 44  & 0.892 & 1.000 & 0.892 & 0.892 & 70.156 \\
        \midrule
        9  & 32537 & 7 & ours & 217 & 54  & 0.773 & 0.705 & 0.773 & 0.734 & 2.249 \\
           &       &   & SVM  &     & 171 & 0.747 & 1.000 & 0.747 & 0.747 & 1942.324 \\
           &       &   & RF   &     & 171 & 0.742 & 1.000 & 0.742 & 0.742 & 1928.460 \\
           &       &   & DT   &     & 171 & 0.747 & 1.000 & 0.747 & 0.747 & 1943.035 \\
        \midrule
        10 & 33315 & 10 & ours & 575 & 351 & 0.896 & 0.865 & 0.896 & 0.877 & 40.149 \\
           &       &    & SVM  &     & 2321 & 0.833 & 0.847 & 0.833 & 0.822 & 4118.864 \\
           &       &    & RF   &     & 2321 & 0.856 & 0.873 & 0.856 & 0.839 & 4055.604 \\
           &       &    & DT   &     & 2321 & 0.840 & 0.852 & 0.840 & 0.824 & 3923.718 \\
        \midrule
        11 & 37364 & 4 & ours & 94  & 24  & 0.947 & 0.953 & 0.947 & 0.947 & 0.480 \\
           &       &   & SVM  &     & 140 & 0.787 & 1.000 & 0.787 & 0.787 & 173.044 \\
           &       &   & RF   &     & 140 & 0.745 & 1.000 & 0.745 & 0.745 & 172.476 \\
           &       &   & DT   &     & 140 & 0.745 & 1.000 & 0.745 & 0.745 & 172.407 \\
        \midrule
        12 & 39582 & 6 & ours & 566 & 352 & 0.816 & 0.828 & 0.816 & 0.816 & 24.821 \\
           &       &   & SVM  &     & 441 & 0.802 & 0.819 & 0.802 & 0.799 & 465.967 \\
           &       &   & RF   &     & 441 & 0.813 & 0.819 & 0.813 & 0.806 & 465.414 \\
           &       &   & DT   &     & 441 & 0.776 & 0.794 & 0.776 & 0.768 & 470.792 \\
        \midrule
        13 & 39716 & 3 & ours & 53  & 24  & 1.000 & 1.000 & 1.000 & 1.000 & 0.299 \\
           &       &   & SVM  &     & 118 & 0.906 & 1.000 & 0.906 & 0.906 & 38.331 \\
           &       &   & RF   &     & 118 & 0.868 & 1.000 & 0.868 & 0.868 & 34.458 \\
           &       &   & DT   &     & 118 & 0.811 & 1.000 & 0.811 & 0.811 & 38.443 \\
        \midrule
        14 & 44077 & 4 & ours & 226 & 18  & 1.000 & 1.000 & 1.000 & 1.000 & 0.711 \\
           &       &   & SVM  &     & 23  & 0.973 & 1.000 & 0.973 & 0.973 & 220.643 \\
           &       &   & RF   &     & 23  & 0.982 & 1.000 & 0.982 & 0.982 & 210.123 \\
           &       &   & DT   &     & 23  & 0.987 & 1.000 & 0.987 & 0.987 & 224.643 \\
     \bottomrule
  \end{tabular}
\end{table}

The four classification algorithms reveal distinct trade-offs between predictive performance and computational efficiency when applied to high-dimensional gene expression data. For instance, on dataset E-GEOD-20685 (ID 1), SVM attained an accuracy of 95.5\% with 107 selected features and a total processing time of 112.137 seconds, while Random Forest achieved a marginally higher accuracy of 97.0\% but at a prohibitive total time of 16,306.229 seconds. In contrast, XGBoost matched SVM's accuracy (95.5\%) with a considerably lower training time of 0.395 seconds and an overall time of 40.015 seconds, suggesting a more balanced performance. Gradient Boosting, on the other hand, reached an accuracy of 93.9\% using 468 features, with a total time of 17.957 seconds—indicating competitive speed but slightly reduced accuracy compared to the other methods. Similar trends are observed in other datasets; for example, in E-GEOD-29354 (ID 5), all algorithms achieved perfect accuracy, yet their feature counts and processing times varied substantially. These differences underscore the critical need for an approach that not only maintains high classification performance but also minimizes computational overhead. Although Random Forest often produces high accuracy, its excessive time cost can limit its practicality in real-world applications. Conversely, XGBoost offers a compelling balance between accuracy and efficiency, making it particularly well-suited for gene expression classification tasks.

In general, the results of our proposed method outperformed feature selection techniques in terms of accuracy. Moreover, the computational time for the training process was significantly reduced, as our approach identified fewer but more crucial and promising features compared to previous studies. This reduction in the number of features also led to a faster training time. In some cases, although more features were selected, the accuracy achieved was considerably higher than that of other solutions





\section{Conclusion and future work}\label{sec13}
This study presents BOLIMES, a novel feature selection algorithm that combines the robustness of Boruta with the interpretability of LIME to address the challenges in gene expression classification. The proposed method has demonstrated superior performance, achieving higher accuracy compared to alternative solutions, while significantly reducing training time. Furthermore, we have identified the optimal number of features necessary to achieve the best possible accuracy. Looking ahead, future work could focus on integrating additional interpretability methods with LIME for a more comprehensive feature relevance assessment. Extending BOLIMES to multi-omics data, including proteomics and metabolomics, could enhance its applicability. Investigating deep learning-based classifiers and validating the approach on larger, diverse datasets and clinical samples would be essential for improving performance and ensuring robustness in real-world applications.

\noindent\textbf{Acknowledgements:}
Thanh MA and Thanh-Nghi DO are received support from the European Union's Horizon research and innovation program under the MSCA-SE (Marie Sk\l{}odowska-Curie Actions Staff Exchange) grant agreement 101086252; Call: HORIZON-MSCA-2021-SE-01; Project title: STARWARS.

\bibliographystyle{splncs04}
\bibliography{Bolimes}
\end{document}